\RequirePackage{amsmath}
\documentclass[runningheads]{llncs} 
\usepackage{xcolor}
\usepackage{footnote}
\usepackage{graphicx,multicol,footmisc}
\usepackage{pgf}
\usepackage{algorithm,algorithmic}
\usepackage{support-caption}
\usepackage{caption,subcaption}
\usepackage{mathtools} 
\usepackage{hyperref}
\usepackage{url}

\usepackage{tikz}

\usetikzlibrary{positioning, decorations.pathmorphing,arrows,arrows.meta,arrows, backgrounds,3d,decorations.text,shapes.arrows,positioning,fit}

\begin{document}
\mainmatter              % start of a contribution
\title{Semi-Supervised Learning using Siamese Networks}
\titlerunning{SSL using Siamese Networks}  % abbreviated title (for running head)
\author{Attaullah Sahito,  Eibe Frank, \and Bernhard Pfahringer %\ \inst{1}
}
\authorrunning{Attaullah et al.} % abbreviated author list (for running head)
\tocauthor{Attaullah Sahito, Eibe Frank, and Bernhard Pfahringer
	%	, 
}
\institute{Department of Computer Science, University of Waikato, Hamilton, New Zealand.\\
\email{a19@students.waikato.ac.nz}, \email{\{eibe,bernhard\}@waikato.ac.nz}
%\texttt{http://cs.waikato.ac.nz/\homedir a19} 
}

\maketitle    

\begin{abstract}
Neural networks have been successfully used as classification models yielding state-of-the-art results when trained on a large number of labeled samples. These models, however, are more difficult to train successfully for semi-supervised problems where small amounts of labeled instances are available along with a large number of unlabeled instances. This work explores a new training method for semi-supervised learning that is based on similarity function learning using a Siamese network to obtain a suitable embedding. The learned representations are discriminative in Euclidean space, and hence can be used for labeling unlabeled instances using a nearest-neighbor classifier. Confident predictions of unlabeled instances are used as true labels for retraining the Siamese network on the expanded training set. This process is applied iteratively. We perform an empirical study of this iterative self-training algorithm. For improving unlabeled predictions, local learning with global consistency \cite{zhou2004learning} is also evaluated. 

\end{abstract}

\section{Introduction}
The modern world generates vast amounts of data and provides many opportunities to exploit it. However, frequently this data is complex, noisy, and lacks obvious structure. Therefore, explicit modeling of, for example, its distribution is too challenging for a human agent. On the other hand, a human can specify an explicit procedure, i.e., an algorithm, for how to construct such a model. Machine learning (ML) is concerned with algorithms that enable computers to learn from data in this way, especially algorithms for prediction. Many ML algorithms need labeled data for such a task, but it is common that fewer labeled data are available than unlabeled ones. Manual labeling is costly and time-consuming. Hence, there is an ever-growing need for ML methods to work with a limited amount of labeled data and also make efficient use of the side information available from unlabeled data. Algorithms designed to do so are known as semi-supervised learning algorithms.

%\subsection{Semi-Supervised Learning}
Supervised learning algorithms employ labeled data to predict class labels for unlabeled examples accurately. Unsupervised learning algorithms search for structure in data, which can then be used as a heuristic to infer labels for these examples, on the basis of assumptions about the structure of data. Semi-Supervised learning (SSL) algorithms lie somewhere between supervised and unsupervised learning. SSL methods are designed to  work with labeled $L=\{(x_1,y_1),(x_2,y_2),...,(x_{|L|},y_{|L|})\}$ and unlabeled instances $U= \{{x^{'}_1},{x^{'}_2},...,{x^{'}_{|U|}} \}$, where $X$ and $Y$ relate to an input space and  output space, $x_i, x^{'}_{j} \in X (i= 1,2,...,|L| , j= 1,2,...,|U|)$ are examples and $y_i \in Y$ are labels of $x_i$ and $Y=\{1,2,3,...,c\}$, $c$ being the number of classes. Usually, these methods assume a much smaller number of labeled instances than unlabeled ones i.e., $|L|\ll|U|$, because unlabeled instances are more useful when we have a few labeled instances. SSL has proven to be useful especially when we are dealing with anti-causal or confounded problems \cite{peters2017elements}.   

Without making any assumptions on how the inputs and outputs are related it is impossible to justify semi-supervised learning as a principled approach \cite{chapelle2006semi}. Like the authors in that paper, we make the same three assumptions: 
\begin{enumerate}
    \item If two points $x_1 , x_2$ are close in a high-density region, then their corresponding outputs $y_1 , y_2$ should also be close.
    \item If points are in the same structure (referred to as cluster or  manifold), they are likely to be of the same class.
    \item The decision boundary between classes should lie in a low-density region of input space.
\end{enumerate} 

In this work, we will consider a new training method designed to be used with deep neural networks in the semi-supervised learning setting. Instead of the usual approach of learning a direct classification model based on cross-entropy loss, we will use the labeled examples for learning a similarity function between instances, such that instances of the same class are considered similar and those instances belonging to different classes are considered dissimilar. Under this similarity function, which is parameterized by a neural network, the features (embeddings) of labeled examples will be grouped together according to the class labels, in Euclidean space. In addition, we will use these learned embeddings to assign class labels to unlabeled examples. We do this using a simple nearest-neighbor classifier. Following that, confident predictions for unlabeled instances are added to the labeled examples for  retraining of the neural network iteratively. In this way, we are able to achieve significant performance improvements over supervised-only training.

\section{Related Work}
Semi-supervised learning has been under study since the 1970s \cite{mclachlan1975iterative}. Expectation-Maximization (EM) \cite{nigam2006semi} works by labeling unlabeled instances with the current supervised model's best prediction in an iterative fashion (self-learning), thereby providing more training instances for the supervised learning algorithm. Co-training \cite{blum1998combining} is a similar approach, where two models are trained on two separate subsets of the data features. Confident predictions from one model are then used as labeled data for the other model. Co-EM \cite{brefeld2004co} combines co-training with EM and achieved better results than either of them. Another, graph-based SSL method, LLGC (Local Learning with Global Consistency) \cite{zhou2004learning}, works by propagating labels from labeled to unlabeled instances until labels are stable, maintaining local and global consistency.

There is a substantial amount of literature available on SSL techniques using deep neural network based on autoencoders \cite{rasmus2015semi,maaloe2016auxiliary}, generative adversarial networks (GAN)  \cite{salimans2016improved,dai2017good,wei2018improving} and based on regularization \cite{laine2016temporal,sajjadi2016regularization,miyato2017virtual}. The Pseudolabel \cite{lee2013pseudo} approach is a deep learning version of self-learning with an extra loss from  regularization and the reconstruction of a denoising autoencoder.

Our method builds on work investigating similarity metric learning using neural networks.  \cite{chopra2005learning} used a network with the contrastive loss for face verification in a supervised fashion. \cite{schroff2015facenet} suggested network training to be based on triplets of examples. This work was extended to the semi-supervised paradigm \cite{weston2012deep} for the image classification task. \cite{hoffer2016semi} tries to minimize the sum of cross-entropy and ratio loss between class indicators (sampled from labeled examples for each class) and the intra-class distances of instances calculated based on embeddings.

We train our network based on triplets of images and  use the triplet margin loss \cite{schroff2015facenet}. We found this to perform better than the contrastive loss or the ratio loss in our experiments, while the network is trained in a  self-learning fashion. For improving intermediate predictions, we use LLGC \cite{zhou2004learning} in  order to get better labels for unlabeled instances in subsequent iterations.  Although triplet networks and LLGC are not new, this is the first attempt, to our knowledge, of combining these two approaches for semi-supervised learning.

 \section{Siamese Networks}
\begin{figure}%[htbp]
	\centering
	\includegraphics[scale=0.830] {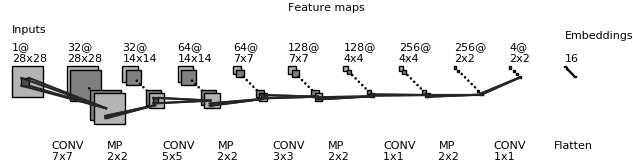}  
	\caption{Network Architecture}
	\label{fig:siamese}
\end{figure}

Siamese networks \cite{bromley1993signature} are neural networks that are particularly efficient when we have a large number of classes and a few labeled instances per class. Siamese networks can be thought of multiple networks with identical copies of the same function, with the same weights. They can be  employed for training a similarity function given labeled data. Fig. \ref{fig:siamese} shows a simple network architecture based on convolutional (CONV) and max-pooling (MP) layers. An input example is passed to the network for computing the embeddings.
Different losses are used for training Siamese networks, such as contrastive loss, margin-based loss, and triplet loss. Network parameters are updated according to the loss calculated on embeddings.
\subsection{Triplet Loss}
The triplet loss \cite{schroff2015facenet} has been used for face recognition. A triplet's anchor example $ a $, positive example $ p $, and negative example $ n $  are provided as a training example to the network for getting corresponding embeddings. During optimisation of the network parameters, we draw all possible triplets from labeled examples based on class labels. For each mini-batch used in stochastic gradient descent, all valid triplets$(i,j,k)$ are selected where $labels[i]=labels[j], i \ne j $ and $labels[i] \ne labels[k]$. Then the loss is calculated according to the following equation using the Euclidean distance $d(.,.)$ between the embedded examples:
\begin{equation}
\label{loss}
\mathcal{L} = max(d(a, p) - d(a, n) + m, 0)
\end{equation}
where $ m $ is the so-called "margin" and constitutes a hyperparameter.

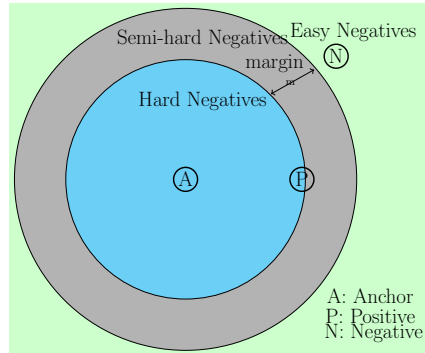
\begin{figure}%[!htbp]
	\centering
	\scalebox{0.455}{		\begin{tikzpicture}	[background rectangle/.style={fill=green!20},
	show background rectangle]

	\draw [fill=gray!60] (0,0) circle (5.0cm) {};
	\draw [fill=cyan!50](0,0) circle (3.5cm) {};
	\draw [very thick, minimum size=7mm] (0,0) node[circle,inner sep=1pt,draw] {\LARGE A};

	\draw [very thick, minimum size=7mm]	(3.4,0) node[circle,inner sep=0.5pt,draw] {\LARGE P};
	\draw [very thick, minimum size=7mm]	(4.4,3.6) node[circle,inner sep=0.5pt,draw] {\LARGE N};

	\node[->] at (2.6,3.4) { \LARGE margin};
	\node[->] at (4.9,4.3) {\LARGE Easy Negatives};
	\node[->] at (0.5,4.1) {\LARGE Semi-hard Negatives};
	\node[->] at (0.5,2.3) {\LARGE Hard Negatives};

	\draw [<->] (2.5,2.5) -- ++(1.25,20pt) node[pos=0.45]{m};
    
    \node[->] at (5.4,-3.4) {\LARGE A: Anchor};
    \node[->] at (5.4,-4) {\LARGE P: Positive};
    \node[->] at (5.5,-4.5) {\LARGE N: Negative};
	
	\end{tikzpicture}} 
	\caption{Triplet loss}
	\label{fig:triplet-loss}
\end{figure}

As illustrated in Fig. \ref{fig:triplet-loss}, the triplet loss attempts to push away the embedded negative example $ n $ from the embedded anchor example $ a $ based on a given margin $ m $ and the given positive example $ p $. Depending on the location of the negative example with respect to the anchor and the positive example, it is possible to distinguish between hard negative examples, semi-hard negative examples, and easy negative examples. The latter are effectively ignored during optimisation because they yield the value zero for the loss.
\subsection{Self-learning using Siamese networks}
In the first iteration of our semi-supervised learning approach, to be able to label (some of) the unlabeled examples instances, the Siamese network is trained on labeled examples only, using triplet loss. Then the standard nearest neighbor classifier is used to predict labels for the unlabeled examples and a fixed percentage $ p $ of unlabeled examples is chosen based on their distance to the labeled instances and added to the set of labeled examples for the next iteration. Throughout, embedded data is used to calculate distances. For more details see the pseudo-code in Listing \ref{alg:siamese}.
\begin{algorithm}%[H]
	\begin{algorithmic}[1]
		\STATE\textbf{Input:} Labeled examples ($x_L,y_L$), Unlabeled examples $x_U$, number of meta-iterations $i$ and  selection percentage $p$
		\FOR {1 to $i$}
		\STATE $train\_siamesenetwork(x_L,y_L)$ 
		\STATE $embed_U = siamesenetwork({x_U})$
		\STATE $embed_L = siamesenetwork({x_L})$
		\STATE $labels_U,dist_U = KNN(embed_U,embed_L,y_L)$
		\STATE $sorted\_dist_U,sorted\_labels_U = sort(dist_U, labels_U)$
		\STATE $x_{new},y_{new} = select\_top(sorted\_dist_U,sorted\_labels_U,p)$ 
		\STATE $x_L,y_L =  concat((x_L,y_L),(x_{new},y_{new}))$
		\STATE $x_U = delete\_from(x_U, x_{new})$
		\ENDFOR
	\end{algorithmic}
	\caption{Proposed approach based on Siamese self-training }
	\label{alg:siamese}
\end{algorithm}
\section{Local Learning with Global Consistency (LLGC)}
We also investigate local learning with global consistency \cite{zhou2004learning} in addition to the nearest-neighbor classifier. LLGC works by propagating label information to the neighbors of an example. The goal of LLGC is to predict labels for unlabeled instances. The algorithm initializes a  matrix $ Y_{n \times c} $  to represent label information, where $Y_{ij}=1$ if example $i$ is labeled as $j$, and otherwise $Y_{ij}=0$. We implement a little variation here for the unlabeled examples: instead of using $Y_{ij}=0$ for all $j$ when $i$ is unlabeled, we use predicted labels obtained with the nearest-neighbour classifier after training the Siamese network. 

LLGC is based on calculating an adjacency matrix. This adjacency matrix is then used to establish a matrix $S$ that is applied to update the label probabilities for the unlabeled examples.
The adjacency matrix is calculated using Eq. \ref{eq:w-calcultion} by employing embeddings $f(x_i)$ and $f(x_j)$ for each pair of two examples $x_i$ and $x_j$, obtained from the Siamese network. The parameter $\sigma$ is a hyper-parameter.  
\begin{equation}
 W_{ij}= 
\begin{cases}
e^{-\sigma \times |f(x_i)-f(x_j)|^2} ,   &\text{if $i \ne j$} \\
0 &\text{if $i = j $}.
\end{cases}
\label{eq:w-calcultion}
\end{equation}
The matrix $S$ is computed as: 
\begin{equation}
S = D^{-1/2} \times W \times D^{-1/2}
\end{equation}
where $D$ is a diagonal matrix: $D_i = \sum_{j=1}^{n} W_{ij}$.
The initial matrix of label probabilities is set to  $ F(0) = Y $, and the probabilities are updated by:
\begin{equation}
F(t+1) = S.F(t)\times \alpha + (1-\alpha)\times Y
\end{equation}
where $ \alpha \in [{0,1})$ is a hyper-parameter for controlling the propagation of label information. The above operation is repeated till convergence. Finally, labels for the unlabeled instances are calculated as:
\begin{equation}
 y_i= \operatorname*{argmax}_{j\leq c} F_{ij}
\end{equation}

For efficiently using unlabeled instances, the Siamese network is first trained on labeled examples only, using triplet loss. Then the nearest-neighbor classifier is used to predict labels for unlabeled examples. Then, following that, labeled and unlabeled embeddings along with labels are passed to LLGC. After a certain number of iterations of LLGC, a fixed percentage $ p $ of unlabeled examples are chosen based on their LLGC score and added to the  labeled examples for the next iteration. For more details see the pseudo-code in Listing \ref{alg:llgc}.

\begin{algorithm}%[H]
	\begin{algorithmic}[1]
	\STATE\textbf{Input:} Labeled examples ($x_L,y_L$), Unlabeled examples $x_U$, number of meta-iterations $i$, selection percentage $p$, $\alpha$ and $\sigma$ parameters for LLGC.   
	\FOR {1 to $i$} 
		\STATE $train\_siamesenetwork(x_L,y_L)$ 
		\STATE $embed_U = siamesenetwork({x_U})$
		\STATE $embed_L = siamesenetwork({x_L})$
		\STATE $labels_U = KNN(embed_U,embed_L,y_L)$
		\STATE $LLGC\_labels,LLGC\_score = LLGC(embed_{L},embed_{U},[y_L,labels_U],\sigma,\alpha)$
		\STATE $labels_U = LLGC\_labels[len(x_L):]$ 
		\STATE $x_{new},y_{new} = select\_top(LLGC\_score,p,x_{U},labels_U)$ 
		\STATE $x_L,y_L =  concat((x_L,y_L),(x_{new},y_{new}))$
		\STATE $x_U = delete\_from(x_U, x_{new})$
	\ENDFOR
	\end{algorithmic}
	\caption{Proposed approach based on LLGC self-training}
	\label{alg:llgc}
\end{algorithm}

\section{Experiments}
We consider four standard image classification problems for our evaluation. For all experiments, a small subset of labeled examples was chosen according to standard semi-supervised learning practice, with a balanced number of examples from each class, and the rest were considered as unlabeled. Final accuracy was calculated on the standard test split for each dataset. No data augmentation was applied to the training sets. Siamese networks were trained using triplet loss with margin $ m =0.3$ for all datasets. 

A simple convolutional network architecture\footnote{Source code available at \url{https://github.com/attaullah/Self-training/blob/master/Metric_learning.md}} was chosen for each dataset to ensure performance achieved was due to the proposed method and not the network architecture. For more details about the network architectures, see Table \ref{table:network}. Layer descriptions use (feature-maps, kernel-size, stride, padding) for convolutional layers and (pool-size, stride) for pooling layers. The simple model is used for MNIST, Fashion MNIST, and SVHN, and produces 16-dimensional embeddings, while the  CIFAR-10 model produces 64-dimensional embeddings. We trained the networks using mini-batch sizes 50, 100, and 200. We found that batch size 50 was insufficient and 200 did not yield significant improvements compared to batch size 100. Batch size = 100 is used for all experiments, with Adam \cite{kingma2014adam} as the optimizer for updating network parameters for 200 epochs. Our proposed approaches Siamese self-training (Algorithm \ref{alg:siamese}) and LLGC self-training (Algorithm \ref{alg:llgc}) respectively were run for 25 meta-iterations. For LLGC, $\alpha=0.99$ is used in all experiments, while $\sigma$ is optimized for each dataset. The final test accuracy is computed using a k-NN classifier with $k=1$ for simplicity. Our results were averaged over 3 random runs, using a different random initialization of the Siamese network parameters for each run and random selection of initially labeled examples except SVHN. We set a baseline by (a) training the network on the small number of the labeled instances only, and by (b) using all the labeled instances. These two baselines should provide good empirical lower and upper bounds for the semi-supervised error rates. 

\begin{table}%[!ht] % float table to adjust text instead of just two tables on whole page...
    \caption{Network Model}
    \label{table:network}
    \begin{center}
    \begin{tabular}{|l|l|}
        \hline
                    Simple(\#parameters=163908) &  CIFAR-10(\#parameters=693792)\\
        \hline
                    INPUT  &     INPUT \\ 
        \hline
                    Conv-Relu(32,7,1,2)  &     Conv-Relu-BN(192,5,1,2) \\
        			Max-Pooling(2,2)  &   Conv-Relu-BN(160,1,1,2) \\
        			Conv-Relu(64,5,1,2)   &   Conv-Relu-BN(96,1,1,2) \\
        			Max-Pooling(2,2)  &  Max-Pooling(3,2) \\
        			Conv-Relu(128,3,1,2)  & Conv-Relu-BN(96,5,1,2) \\
        			Max-Pooling(2,2)  &   Conv-Relu-BN(192,1,1,2) \\
        			Conv-Relu(256,1,1,2)  &  Conv-Relu-BN(192,1,1,2) \\
        			Max-Pooling(2,2)  &   Max-Pooling(3,2) \\
        			Conv(4,1,1,2)  &   Conv-Relu-BN(192,3,1,2) \\
        			 Flatten() &  Conv-Relu-BN(64,1,1,2) \\
        			                &    Avg-Pooling(8,1) \\
        \hline
    \end{tabular}
    \end{center}
    %	\vspace{-10mm}   % for reducing too much white space after the table.
\end{table}

We now consider the datasets used in our experiments. The MNIST dataset consists of gray-scale 28 by 28 images of handwritten digits. We select only 100 instances (10 from each class) as labeled instances initially. We apply our algorithms with a selection percentage $p=10\%$  and the LLGC-based method with $\sigma=1.8$. Table \ref{tabel:mnist} shows noticeable improvements over the supervised-only approach when compared with the proposed semi-supervised approaches, when using the same number of labeled examples.
\begin{table}%[!ht]
	\caption{MNIST  Test error \%.}
	\label{tabel:mnist}
\begin{center}
\begin{tabular}{|l|l|l|}
\hline
   \# labels& 100-Labeled  & All (60000) \\
\hline
		Supervised-only		&    $9.73\pm0.74$      & $0.6\pm0.04$         \\
		Siamese self-training	&  $\mathbf {3.24\pm0.32}$      &      \quad  \quad --    \\
		LLGC self-training		& $3.50\pm0.14$       &   \quad  \quad --  \\   
\hline
\end{tabular}
\end{center}
\end{table}

The Fashion MNIST dataset consists of 28 by 28 gray-scale images showing fashion items. 100 instances are considered as labeled initially. Again, we use selection percentage $p=10\%$ and $\sigma=3.2$. Table \ref{tabel:fmnist} again shows noticeable improvement over the supervised-only approach when compared with the proposed semi-supervised approaches, when using the same amount of labeled data.
\begin{table}%[h]
	\caption{Fashion MNIST  Test error \%.}
	\label{tabel:fmnist}
	\begin{center}
	\begin{tabular}{|l|l|l|}
		\hline
			\# labels& 100-Labeled  & All (60000) \\
		\hline
		Supervised-only		&    $26.72\pm 1.23$      & $9.66\pm 0.10$         \\
		Siamese self-training	&    $23.33\pm 0.43$       & \quad  \quad --  \\  
		LLGC self-training	&    $\mathbf{23.23\pm 0.67}$       & \quad  \quad --  \\   
		\hline
	\end{tabular}
\end{center}
\end{table}

SVHN comprises 32x32 RGB images of house numbers, taken from the Street View House Numbers dataset. Each image can have  multiple digits, but only the digit in the center is considered for prediction. The proposed approaches are evaluated using 1000 labeled instances initially, with selection percentage $p=5\%$, and $\sigma=2.4$. Table \ref{tabel:svhn} shows noticeable improvement over the supervised-only approach when compared to the proposed approaches when 1000 labeled examples are used. Interestingly, purely Siamese self-training again performs better than LLGC self-training in this case. 
\begin{table}%[htpb]
	\caption{SVHN   Test error \%.}
	\label{tabel:svhn}
	\begin{center}
		\begin{tabular}{|l|l|l|}
		\hline
			\# labels& 1000-Labeled  & All (73275) \\
		\hline
		Supervised-only		&    $30.33\pm1.55$      & $12.26\pm0.52$         \\
		Siamese self-training	&    $\mathbf{20.09\pm 3.22}$      & \quad  \quad --  \\  
		LLGC self-training		&    $27.23\pm 0.99$       &   \quad  \quad --  \\   
		\hline
	\end{tabular}
\end{center}
\end{table}

The CIFAR-10 dataset contains 32 by 32 RGB images of ten classes. The proposed semi-supervised approaches are evaluated using 4000 labeled instances initially, with selection percentage $p=5\%$, and $\sigma=2.4$.  Table \ref{tabel:cifar10} shows little improvement over the supervised-only approach when compared to the proposed semi-supervised approaches. Siamese self-training performs better than LLGC self-training.
\begin{table}%[htpb]
	\caption{CIFAR-10  Test error \%.}
	\label{tabel:cifar10}
	\begin{center}
	\begin{tabular}{|l|l|l|}
		\hline
			\# labels & 4000-Labeled  & All (50000) \\
		\hline
		Supervised-only		&    $40.87\pm 0.56$      & $21.51\pm 0.88 $         \\
		Siamese self-training		&   $\mathbf{36.56\pm 0.74}$      &   \quad  \quad --  \\  
		LLGC self-training		&    $40.06\pm 0.62$       &   \quad  \quad --  \\  
		\hline
	\end{tabular}
\end{center}
\end{table}

\begin{figure}
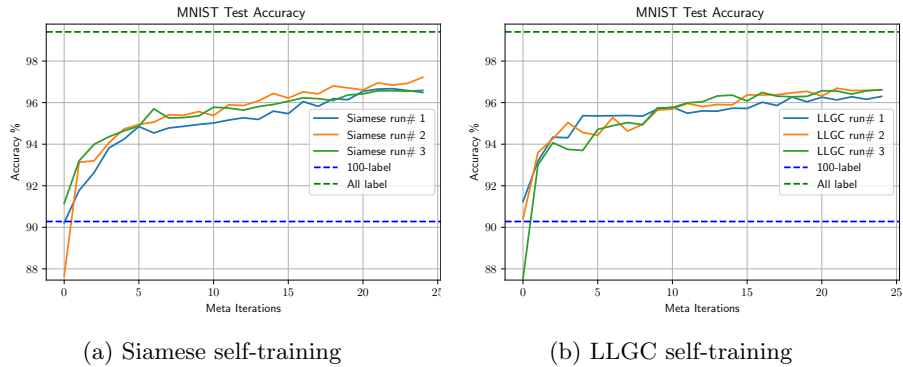
 %[!htpb]
	\centering 
	\begin{subfigure}{0.5\textwidth}
		\scalebox{0.444} {\input{imgs/mnist_siamese.pgf} } 
		\caption{Siamese self-training}
		\label{fig:1}
	\end{subfigure}%\hfil 
	\begin{subfigure}{0.5\textwidth}
		\scalebox{0.444} {\input{imgs/mnist_llgc.pgf} } 
		\caption{ LLGC self-training}
		\label{fig:2}
	\end{subfigure} 
	\caption{MNIST-100 Comparison of Siamese self-training vs. LLGC self-training.}
	\label{fig:mnist-comparison}
	\end{figure}

\begin{figure}
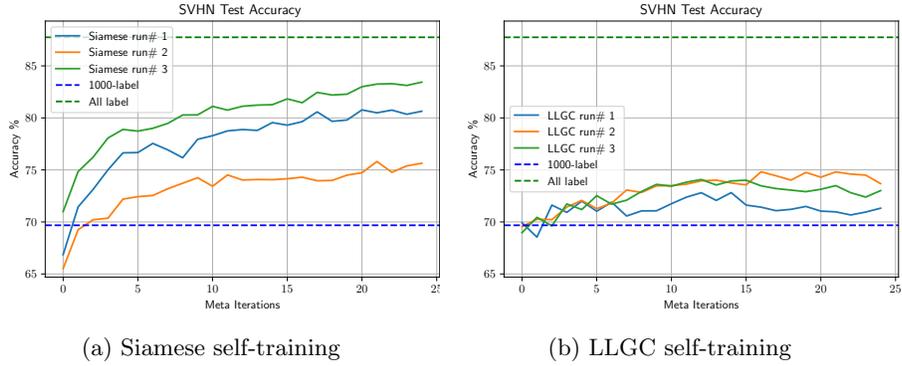
 %[!htpb]
	\centering 
	%\\medskip
	\begin{subfigure}{0.5\textwidth}
		\scalebox{0.444} {\input{imgs/svhn_siamese.pgf} } 
		\caption{Siamese self-training}
		\label{fig:5}
	\end{subfigure}%\hfil 
	\begin{subfigure}{0.5\textwidth}
			\scalebox{0.444} {\input{imgs/svhn_llgc.pgf} } 
		\caption{LLGC self-training}
		\label{fig:6}
	\end{subfigure}
	
	\caption{SVHN-1000 Comparison of Siamese self-training vs. LLGC self-training.}
	\label{fig:svhn-comparison}
\end{figure}

\begin{figure}
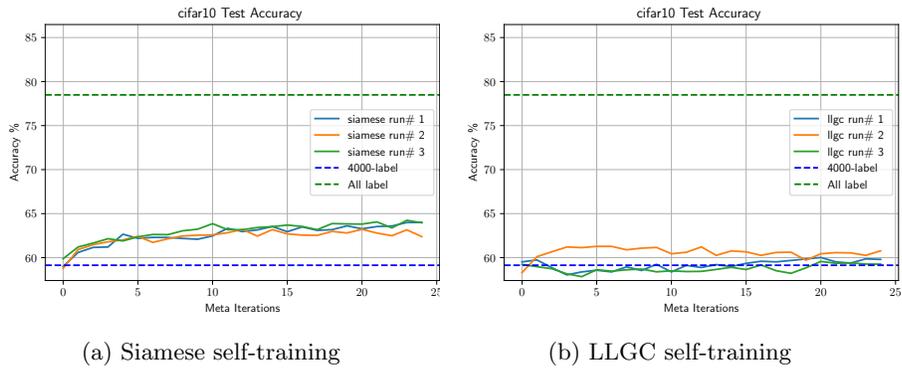
 %[!htpb]
	\centering 
	%\	\medskip
	\begin{subfigure}{0.5\textwidth}
			\scalebox{0.444} {\input{imgs/cifar10_siamese.pgf} } 
		\caption{Siamese self-training}
		\label{fig:7}
	\end{subfigure}%\hfil 
	\begin{subfigure}{0.5\textwidth}
			\scalebox{0.444} {\input{imgs/cifar10_llgc.pgf} } 
		\caption{LLGC self-training}
		\label{fig:8}
	\end{subfigure}
	\caption{ CIFAR10-4000 Comparison of Siamese self-training vs. LLGC self-training.}
	\label{fig:cifar10-comparison}
\end{figure}

Figures \ref{fig:mnist-comparison}, \ref{fig:fmnist-comparison}, \ref{fig:svhn-comparison} and  \ref{fig:cifar10-comparison} show a detailed comparison between Siamese self-training and LLGC self-training across three different runs of all four datasets; MNIST, Fashion MNIST, SVHN, and CIFAR-10.  The accuracy curves show definite improvement with respect to the supervised-only version on all datasets using Siamese self-training as well as LLGC self-training. However, CIFAR-10 and SVHN seem to get low or negligible additional improvement from LLGC self-training compared to Siamese self-training only.
	
\begin{figure} %[!htpb]
	\centering 
	%\medskip
	\begin{subfigure}{0.5\textwidth}
		\scalebox{0.444} {\input{imgs/fmnist_siamese.pgf} } 
		\caption{Siamese self-training}
		\label{fig:3}
	\end{subfigure}%\hfil 
	\begin{subfigure}{0.5\textwidth}
		\scalebox{0.444} {\input{imgs/fmnist_llgc.pgf} }  
		\caption{ LLGC self-training}
		\label{fig:4} 
	\end{subfigure}
	
	\caption{Fashion MNIST-100 Comparison of Siamese self-training vs. LLGC self-training.}
	\label{fig:fmnist-comparison}
\end{figure}

\begin{figure}
	\centering 
	\begin{subfigure}{0.5\textwidth}
		\includegraphics[scale=0.40] {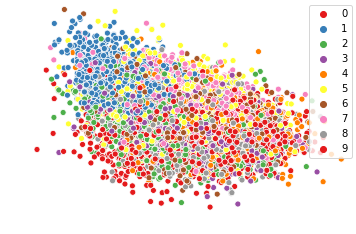}  
		\caption{Before training}
		\label{fig:before}
	\end{subfigure}\hfil 
	\begin{subfigure}{0.5\textwidth}
		\includegraphics[scale=0.44] {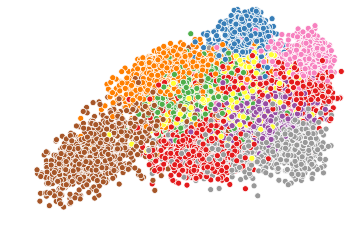}  
		\caption{After training}
		\label{fig:after}
	\end{subfigure} 	
	\caption{MNIST-100: visualisation of 2-dimensional embeddings}
\label{fig:embeddings}
	\vspace{-3.22mm} % fit in 12 page limit
\end{figure}

We also tried to visualize the quality of embeddings learned using the proposed method. We trained an additional model by slightly modifying the simple model \ref{table:network}. In order to get a 2-dimensional embedding, two feature-maps are used instead of 4 in the last convolutional layer, followed by average-pooling(2,2) before the final flattening layer. For this purpose, we considered MNIST. Figure \ref{fig:embeddings} (a) depicts the embeddings for test instances marked in color according to their true class after random initialization of the network. Figure \ref{fig:embeddings} (b) depicts the embeddings for test instances after training the Siamese network with only the 100 labeled MNIST instances. It can be seen that the 10000 test examples' embeddings form clusters in Euclidean space after  training of the network according to the class labels; test examples' embeddings are largely scattered randomly throughout the 2D space before the network is trained.

\section{Conclusion}
In this work, we have shown how neural networks can be used to learn in a semi-supervised setting using small sets of labeled data by replacing the classification objective with an objective for learning a similarity function. This objective is compliant with standard techniques of training the deep neural network and requires no modification of the embedding model. For improving the intermediate prediction of unlabeled instances, we evaluated LLGC, but this yielded little additional benefit compared to k-NN classification alone. Using the method in this work, we were able to achieve significant improvement compared to supervised learning only on MNIST,  Fashion MNIST and SVHN, when training on a small subset of labeled examples, but obtained little improvement on CIFAR-10. We speculate that instead of a fixed selection of unlabeled instances from LLGC's predictions, a threshold-based selection based on the LLGC score will be more beneficial for subsequent iterations of  our meta-algorithm. Also, a more robust convolutional model may help the network in learning distinctive embeddings and achieving state-of-the-art results for the semi-supervised setting.

\bibliographystyle{splncs04} 
\bibliography{References}

\begin{thebibliography}{10}
\providecommand{\url}[1]{\texttt{#1}}
\providecommand{\urlprefix}{URL }
\providecommand{\doi}[1]{https://doi.org/#1}

\bibitem{blum1998combining}
Blum, A., Mitchell, T.: Combining labeled and unlabeled data with co-training.
  In: Proceedings of the eleventh annual conference on Computational learning
  theory. pp. 92--100. ACM (1998)

\bibitem{brefeld2004co}
Brefeld, U., Scheffer, T.: Co-em support vector learning. In: Proceedings of
  the twenty-first international conference on Machine learning. p.~16. ACM
  (2004)

\bibitem{bromley1993signature}
Bromley, J., Bentz, J., Bottou, L., Guyon, I., LeCun, Y., Moore, C., Sackinger,
  E., Shah, R.: Signature verification using a “siamese” time delay neural
  network. Int.]. Pattern Recognit. Artzf Intell  \textbf{7} (1993)

\bibitem{chapelle2006semi}
Chapelle, O., Sch{\"o}lkopf, B., Zien, A.: Semi-supervised learning, ser.
  Adaptive computation and machine learning. Cambridge, MA: The MIT Press
  (2006)

\bibitem{chopra2005learning}
Chopra, S., Hadsell, R., LeCun, Y., et~al.: Learning a similarity metric
  discriminatively, with application to face verification. In: CVPR (1). pp.
  539--546 (2005)

\bibitem{dai2017good}
Dai, Z., Yang, Z., Yang, F., Cohen, W.W., Salakhutdinov, R.R.: Good
  semi-supervised learning that requires a bad gan. In: Advances in Neural
  Information Processing Systems. pp. 6513--6523 (2017)

\bibitem{hoffer2016semi}
Hoffer, E., Ailon, N.: Semi-supervised deep learning by metric embedding. arXiv
  preprint arXiv:1611.01449  (2016)

\bibitem{kingma2014adam}
Kingma, D.P., Ba, J.: Adam: A method for stochastic optimization. arXiv
  preprint arXiv:1412.6980  (2014)

\bibitem{laine2016temporal}
Laine, S., Aila, T.: Temporal ensembling for semi-supervised learning. arXiv
  preprint arXiv:1610.02242  (2016)

\bibitem{lee2013pseudo}
Lee, D.H.: Pseudo-label: The simple and efficient semi-supervised learning
  method for deep neural networks. In: Workshop on Challenges in Representation
  Learning, ICML. vol.~3, p.~2 (2013)

\bibitem{maaloe2016auxiliary}
Maal{\o}e, L., S{\o}nderby, C.K., S{\o}nderby, S.K., Winther, O.: Auxiliary
  deep generative models. arXiv preprint arXiv:1602.05473  (2016)

\bibitem{mclachlan1975iterative}
McLachlan, G.J.: Iterative reclassification procedure for constructing an
  asymptotically optimal rule of allocation in discriminant analysis. Journal
  of the American Statistical Association  \textbf{70}(350),  365--369 (1975)

\bibitem{miyato2017virtual}
Miyato, T., Maeda, S.i., Koyama, M., Ishii, S.: Virtual adversarial training: a
  regularization method for supervised and semi-supervised learning. arXiv
  preprint arXiv:1704.03976  (2017)

\bibitem{nigam2006semi}
Nigam, K., McCallum, A., Mitchell, T.: Semi-supervised text classification
  using em. Semi-Supervised Learning pp. 33--56 (2006)

\bibitem{peters2017elements}
Peters, J., Janzing, D., Sch{\"o}lkopf, B.: Elements of causal inference:
  foundations and learning algorithms. MIT Press (2017)

\bibitem{rasmus2015semi}
Rasmus, A., Berglund, M., Honkala, M., Valpola, H., Raiko, T.: Semi-supervised
  learning with ladder networks. In: Advances in Neural Information Processing
  Systems. pp. 3546--3554 (2015)

\bibitem{sajjadi2016regularization}
Sajjadi, M., Javanmardi, M., Tasdizen, T.: Regularization with stochastic
  transformations and perturbations for deep semi-supervised learning. In:
  Advances in Neural Information Processing Systems. pp. 1163--1171 (2016)

\bibitem{salimans2016improved}
Salimans, T., Goodfellow, I., Zaremba, W., Cheung, V., Radford, A., Chen, X.:
  Improved techniques for training gans. In: Advances in Neural Information
  Processing Systems. pp. 2234--2242 (2016)

\bibitem{schroff2015facenet}
Schroff, F., Kalenichenko, D., Philbin, J.: Facenet: A unified embedding for
  face recognition and clustering. In: Proceedings of the IEEE conference on
  computer vision and pattern recognition. pp. 815--823 (2015)

\bibitem{wei2018improving}
Wei, X., Gong, B., Liu, Z., Lu, W., Wang, L.: Improving the improved training
  of wasserstein gans: A consistency term and its dual effect. arXiv preprint
  arXiv:1803.01541  (2018)

\bibitem{weston2012deep}
Weston, J., Ratle, F., Mobahi, H., Collobert, R.: Deep learning via
  semi-supervised embedding. In: Neural Networks: Tricks of the Trade, pp.
  639--655. Springer (2012)

\bibitem{zhou2004learning}
Zhou, D., Bousquet, O., Lal, T.N., Weston, J., Sch{\"o}lkopf, B.: Learning with
  local and global consistency. In: Advances in neural information processing
  systems. pp. 321--328 (2004)

\end{thebibliography}

\end{document}